\documentclass[11pt]{article}

\usepackage[a4paper,margin=2.35cm]{geometry}
\usepackage{amsmath,amssymb}
\usepackage{array,booktabs,longtable}
\usepackage{xcolor}
\usepackage{listings}
\usepackage{tikz}
\usepackage{authblk}
\usetikzlibrary{arrows.meta,positioning}
\usepackage[colorlinks=true,linkcolor=black,citecolor=black,urlcolor=blue!55!black]{hyperref}

\definecolor{accent}{RGB}{37,99,235}
\definecolor{okc}{RGB}{5,150,105}
\definecolor{warnc}{RGB}{217,119,6}
\definecolor{errc}{RGB}{220,38,38}
\definecolor{muted}{RGB}{100,116,139}
\definecolor{codebg}{RGB}{245,247,250}

\newsavebox{\calloutbox}
\newenvironment{callout}[1][accent]{\par\medskip\noindent
  \def\calloutcolor{#1}\begin{lrbox}{\calloutbox}
  \begin{minipage}{\dimexpr\linewidth-2\fboxsep-2\fboxrule-3pt\relax}}
 {\end{minipage}\end{lrbox}
  \noindent\fcolorbox{\calloutcolor}{\calloutcolor!6}{\usebox{\calloutbox}}\par\medskip}

\hypersetup{
  pdftitle={Token Counts Are Not Model Lineage: A Frozen-Threshold Holdout Study of Black-Box LLM API Fingerprinting},
  pdfauthor={Bo Chen}
}

\title{Token Counts Are Not Model Lineage:\\
A Frozen-Threshold Holdout Study of Black-Box LLM API Fingerprinting}
\author[1]{Bo Chen}
\affil[1]{Institute of Computing Technology, Chinese Academy of Sciences}
\date{}

\begin{document}
\maketitle
\begin{center}
\small Code and data: \url{https://github.com/ictchenbo/which-llm}
\end{center}

\begin{abstract}
Black-box model attribution is increasingly relevant when large language models
(LLMs) are served through relay and reseller APIs.  A tempting low-cost signal is
the prompt-token count returned by an OpenAI-compatible endpoint: two models that
share a tokenizer and chat template may produce the same count sequence up to a
fixed offset.  Yet the validity of this signal for broader \emph{model-family}
attribution has received little direct holdout testing.  We conduct a
frozen-threshold study over 24 labeled endpoint pairs, split evenly into a
development set and an untouched holdout set, with three temporal repeats and 30
controlled texts per pair.  We introduce a validity-gated result contract that
distinguishes an observed dissimilarity from an uninformative measurement caused
by missing usage data, rate limits, or endpoint policy.  The resulting
shift-invariant exact-match score perfectly separates the 12 development pairs,
yielding a frozen threshold of 0.725.  On holdout, however, only 6 of 12 pairs are
eligible under the pre-specified three-repeat rule.  Among eligible pairs,
balanced accuracy is 0.75, sensitivity is 0.50 (95\% Wilson interval
0.15--0.85), and specificity is 1.00 (0.342--1.00).  Two same-family pairs---Qwen
3.8 and DeepSeek V4 variants---fall below the frozen threshold.  Across 4,320
formal API calls, every log is replayable, while holdout contains 189 non-200
responses and 157 successful responses without prompt-token usage.  The study
therefore validates token-count consistency as a fingerprint of a shared
\emph{tokenization stack}, but rejects its use as a standalone necessary test for
model-family lineage.
\end{abstract}

\section{Introduction}

Public APIs increasingly mediate access to models whose weights, tokenizer,
system prompt, provider route, and serving configuration are hidden.  This creates
an attribution problem: given two endpoints, can an auditor determine whether they
serve the same model, variants from the same family, or unrelated systems?  The
question matters for licensing, substitution auditing, incident response, and
claims made by relay services.  Prior work studies passive output-space
fingerprints, generated-query fingerprints, embedded ownership marks, adversarial
fingerprints, and behavioral attribution
\cite{YangWu2024,Jin2024,Iourovitski2024,Yang2025,Hu2026,Fang2026,White2026}.

This paper isolates a much cheaper observable: the sequence of
\texttt{usage.prompt\_tokens} values returned for controlled texts.  Token counts
are attractive because they are deterministic under a fixed request envelope,
inexpensive to collect, and less sensitive to answer wording than behavioral
outputs.  They are also easy to over-interpret.  A count includes not only text
tokenization but also hidden chat-template and system-prompt overhead.  Conversely,
two variants in the same named model family need not retain the same tokenizer,
template, provider route, or accounting convention.

The central question is therefore deliberately narrow:

\begin{callout}
\textbf{Research question.} Does a shift-invariant prompt-token-count fingerprint,
calibrated on known endpoint pairs, generalize as a classifier of named model-family
membership on an untouched holdout set?
\end{callout}

The answer in our experiment is \emph{no}.  The signal is useful but is neither a
necessary condition for named family membership nor universally observable through
a relay API.  This negative result is scientifically useful because the development
set alone gives a misleadingly perfect result.

\paragraph{Contributions.}
\begin{enumerate}
  \item We formalize a validity-gated black-box evidence contract with three
  states---informative, uninformative, and error---so that missing usage or blocked
  requests are not silently converted into model dissimilarity.
  \item We define a simple, auditable, shift-invariant count fingerprint and a
  frozen development/holdout protocol with temporal repeats, pre-specified coverage
  gates, and replayable per-call logs.
  \item We report a genuine holdout failure: development balanced accuracy is 1.00,
  but strict holdout coverage is 0.50 and sensitivity among eligible pairs is 0.50.
  The result constrains what token-count evidence can support.
  \item We quantify an operational observability boundary: rate limits, data-policy
  restrictions, and provider responses that omit usage remove half of the holdout
  pairs from strict evaluation.
\end{enumerate}

\section{Related Work and Scope}

\subsection{Black-box ownership verification}

Ownership verification asks whether a suspect model was derived from a protected
model.  ProFLingo generates queries that elicit distinctive responses without
requiring suspect-model internals or modifications to the protected model
\cite{Jin2024}.  Yang and Wu model LLM outputs as model-specific vector spaces and
compare victim and suspect output subspaces \cite{YangWu2024}.  More recent methods
optimize or embed fingerprints for robustness to post-training, quantization, or
model transformations \cite{Hu2026}.  These methods generally assume either a
protected reference model or the ability to construct fingerprints with privileged
access.  Our setting is weaker: neither endpoint is assumed to be owned by the
auditor, and only ordinary API metadata is observed.

\subsection{Passive attribution and API auditing}

Family attribution predicts a model identity or family from black-box behavior.
Hide and Seek uses LLM-generated discriminative prompts and reports family-level
identification across several model families \cite{Iourovitski2024}.  Yang et al.
show that both passive and proactive identification methods have important limits
across 30 LLMs and real-world APIs \cite{Yang2025}.  KBF fingerprints production
endpoints using numerical recall near a model's knowledge boundary
\cite{Fang2026}.  White et al. distinguish base-model attribution from system-prompt
fingerprinting in conversational agents and show that aggregation over many
conversations improves accuracy \cite{White2026}.  These results motivate a careful
separation between a model, a model family, and a deployed agent configuration.

\subsection{What this paper does not claim}

This paper does not propose an ownership watermark, identify stolen weights, or
prove training ancestry.  It evaluates whether one API-level observable is a
reliable proxy for an \emph{operational} same-family label.  The label is assigned
from provider namespaces, public model names, and version/size designations in the
OpenRouter catalog \cite{OpenRouterModels}.  It is not a cryptographic ground truth
about training data or parameter descent.  We return to this limitation in
Section~\ref{sec:limitations}.

\section{Task and Threat Model}

Let endpoints $A$ and $B$ expose an OpenAI-compatible chat-completions interface.
The auditor has no access to weights, tokenizer files, hidden prompts, provider
logs, or routing policy.  For each text stimulus $x_i$, the endpoint may return a
prompt-token count
\[
  c_A(x_i), c_B(x_i) \in \mathbb{N},
\]
or no valid count.  The endpoint may rate-limit requests, omit usage, or reject a
route because of account-level policy.

The operational binary label is
\[
  y(A,B) =
  \begin{cases}
  1, & \text{same named model family or same base model variant},\\
  0, & \text{different named model families}.
  \end{cases}
\]
Same-family examples include different parameter scales, versions, service tiers,
and reasoning/instruction variants.  Negatives are cross-provider, cross-family
pairs.  The task is intentionally more demanding than exact-model matching:
same-family variants may differ materially.

\paragraph{Non-adversarial setting.}
We evaluate naturally deployed endpoints.  Providers are not assumed to actively
spoof token counts.  A malicious gateway could trivially rewrite usage metadata;
therefore success here would not provide adversarial security.

\section{Method}

\subsection{Controlled requests}

We use 30 texts spanning English, Chinese, mixed scripts, source code, emoji,
punctuation, whitespace, structured formats, URLs, and two long repeated-text
cases.  Each endpoint receives the same user message with
\texttt{temperature=0} and \texttt{max\_tokens=1}.  We accept only the integer
\texttt{usage.prompt\_tokens} value.  Character-based estimates are never mixed
with provider counts.  OpenRouter documents that models may tokenize the same text
differently and exposes usage token counts in compatible responses
\cite{OpenRouterModels}.

\subsection{Shift-invariant exact-match fingerprint}

For a repeat with valid paired counts on index set $V$, define offsets
\[
  d_i = c_B(x_i)-c_A(x_i), \qquad i\in V.
\]
Let $m$ be the most frequent offset (ties follow the deterministic order of the
implementation).  The repeat-level fingerprint score is
\begin{equation}
  s(A,B) = \frac{1}{|V|}\sum_{i\in V}\mathbf{1}[d_i=m].
  \label{eq:score}
\end{equation}
This score equals one when all count differences are explained by a fixed overhead,
such as a stable hidden chat-template prefix.  Every text contributes one vote, so
long samples do not dominate the score as they can dominate Pearson correlation.
We report Pearson and rank correlation only as diagnostics, not as classifier
inputs.

\subsection{Validity gates}

A repeat is \emph{informative} only if (i) at least 20 paired counts are valid and
(ii) paired coverage is at least 0.8.  Otherwise its score is \texttt{None}, not
zero.  A pair is eligible for final classification only if all three pre-specified
repeats are informative.  Its pair score is the arithmetic mean of the three repeat
scores.  This strict rule was fixed before holdout evaluation.

Each evaluator result follows the contract
\begin{center}
\small
\begin{tabular}{ll}
\toprule
Field & Meaning\\
\midrule
\texttt{status} & \texttt{informative}, \texttt{uninformative}, or \texttt{error}\\
\texttt{similarity} & score in $[0,1]$ only when informative; otherwise null\\
\texttt{coverage} & paired valid fraction and related diagnostics\\
\texttt{reason} & machine-readable exclusion explanation\\
\texttt{calibrated} & whether a separately frozen calibration is in force\\
\bottomrule
\end{tabular}
\end{center}

\subsection{Threshold calibration}

For development pair scores $\bar{s}_j$, we enumerate observed scores, adjacent
midpoints, and boundary candidates.  We maximize balanced accuracy,
\[
  \mathrm{BA}=\tfrac12(\mathrm{sensitivity}+\mathrm{specificity}),
\]
and average tied maximizing candidates.  This produces threshold $\tau=0.725$.
The calibration artifact is frozen before holdout evaluation with SHA-256 digest
\texttt{d95b0ea9...0307269}.  Holdout predictions use
\[
  \hat{y}(A,B)=\mathbf{1}[\bar{s}(A,B)\ge\tau]
\]
without re-estimating $\tau$.

\section{Experimental Design}

\subsection{Pair matrix}

We construct 24 endpoint pairs and split them before formal collection into 12
development and 12 holdout pairs.  Each split contains six operational positives
and six negatives.  The development set includes GLM, Qwen, Gemma, Ministral, and
DeepSeek same-family controls plus cross-family negatives.  The holdout introduces
unseen version combinations and provider pairs.  Appendix~\ref{app:pairs} lists all
pairs.

The model catalog and availability snapshot were checked on August 30, 2026 through
OpenRouter's Models API, which exposes model identifiers, architecture metadata,
context limits, and pricing \cite{OpenRouterModelsAPI}.  Catalog presence was used
only as an availability screen; it did not guarantee that a particular account
could route to every endpoint.

\subsection{Frozen sequence}

\begin{figure}[htbp]
\centering
\begin{tikzpicture}[font=\small, node distance=0.8cm and 0.65cm,
  box/.style={draw=accent, rounded corners, fill=accent!5, align=center,
    minimum width=3.2cm, minimum height=1.05cm},
  arrow/.style={-{Latex[length=2.2mm]}, thick, draw=muted}]
  \node[box] (pilot) {Engineering pilot\\1 positive + 1 negative};
  \node[box, right=of pilot] (dev) {Development\\12 pairs $\times$ 3 repeats};
  \node[box, right=of dev] (freeze) {Freeze $\tau=0.725$\\hash calibration artifact};
  \node[box, below=of freeze] (hold) {Untouched holdout\\12 pairs $\times$ 3 repeats};
  \node[box, left=of hold] (audit) {One-shot evaluation\\no threshold tuning};
  \draw[arrow] (pilot) -- (dev);
  \draw[arrow] (dev) -- (freeze);
  \draw[arrow] (freeze) -- (hold);
  \draw[arrow] (hold) -- (audit);
\end{tikzpicture}
\caption{Experimental sequence. The pilot checks engineering only; P1--P3 from
earlier exploratory work are treated as development data, never as holdout proof.}
\label{fig:protocol}
\end{figure}
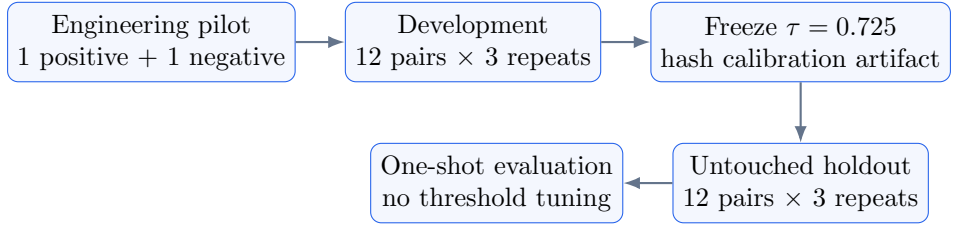

\subsection{Repetition and audit trail}

Each pair is run three times.  Every request and response is logged to an exclusive
JSONL file with a unique run ID, timestamp, redacted authorization header, request
body, full response body, latency, and transport error.  An interrupted repeat
leaves its partial log intact and restarts under an attempt-suffixed run ID; partial
logs are not included in artifacts.  Formal results contain 4,320 calls: 2,160 in
development and 2,160 in holdout.

\section{Results}

\subsection{Development: apparently perfect separation}

All 36 development repeats are informative.  Every formal call returns HTTP 200,
and every response log can be parsed and replayed.  Table~\ref{tab:dev} reports the
three-repeat pair scores.

\begin{table}[htbp]
\centering
\small
\begin{tabular}{p{4.5cm}cccl}
\toprule
Pair & Label & Mean & Std. & Range\\
\midrule
GLM 5.3 Flash / GLM 5.2 & same & 1.0000 & 0 & 1.0000--1.0000\\
Qwen3 235B / Qwen3 30B & same & 0.8889 & 0.1571 & 0.6667--1.0000\\
Qwen3.5 9B / Qwen3.5 35B & same & 0.9445 & 0.0314 & 0.9000--0.9667\\
Gemma4 26B / Gemma4 31B & same & 1.0000 & 0 & 1.0000--1.0000\\
Ministral 3B / Ministral 8B & same & 1.0000 & 0 & 1.0000--1.0000\\
DeepSeek V4 Flash / V4 Pro & same & 0.7889 & 0.0157 & 0.7667--0.8000\\
Six cross-family negatives & different & 0.2667--0.5333 & --- & 0.2000--0.5333\\
\bottomrule
\end{tabular}
\caption{Development pair scores. The clean gap motivates but does not validate the
frozen threshold. Qwen3 shows substantial repeat variability.}
\label{tab:dev}
\end{table}

At $\tau=0.725$, development sensitivity, specificity, and balanced accuracy are
all 1.00.  The Qwen3 pair ranges from 0.6667 to 1.0, already showing that a single
repeat would be unstable even though the three-repeat mean remains above threshold.

\subsection{Holdout: incomplete coverage and false negatives}

Only 6 of 12 holdout pairs satisfy the strict three-repeat eligibility rule.  On
these six, the confusion matrix is TP$=2$, TN$=2$, FP$=0$, FN$=2$.  Balanced
accuracy is 0.75.  Sensitivity is 0.50 with a 95\% Wilson interval of 0.15--0.85;
specificity is 1.00 with interval 0.342--1.00.  The intervals are intentionally
reported because the eligible sample is small.

\begin{figure}[htbp]
\centering
\begin{tikzpicture}[font=\small]
  \def\xzero{4.4}
  \def\scale{8.6}
  \foreach \v in {0,0.25,0.5,0.75,1.0}{
    \draw[gray!20] ({\xzero+\v*\scale},0.25) -- ({\xzero+\v*\scale},-4.65);
    \node[font=\scriptsize,text=muted] at ({\xzero+\v*\scale},-4.9) {\v};
  }
  \draw[warnc,dashed,very thick] ({\xzero+0.725*\scale},0.35) -- ({\xzero+0.725*\scale},-4.65);
  \node[font=\scriptsize,text=warnc,anchor=south] at ({\xzero+0.725*\scale},0.38) {$\tau=0.725$};
  \newcommand{\barrow}[4]{
    \node[anchor=east] at ({\xzero-0.15},#2) {#1};
    \fill[#4] (\xzero,{#2-0.18}) rectangle ({\xzero+#3*\scale},{#2+0.18});
    \node[anchor=west,font=\scriptsize] at ({\xzero+#3*\scale+0.08},#2) {#3};
  }
  \barrow{Qwen3.8 same-family (FN)}{0}{0.5444}{errc}
  \barrow{DeepSeek V4 same-family (FN)}{-0.75}{0.4333}{errc}
  \barrow{Qwen3.6 same-family (TP)}{-1.5}{0.9667}{okc}
  \barrow{Qwen3-VL same-family (TP)}{-2.25}{1.0000}{okc}
  \barrow{GPT / Gemma different (TN)}{-3.0}{0.3000}{accent}
  \barrow{Mistral / DeepSeek different (TN)}{-3.75}{0.2778}{accent}
\end{tikzpicture}
\caption{Eligible holdout pair means under the frozen threshold.  The two false
negatives show that the token-count criterion is not necessary for named family
membership.}
\label{fig:holdout}
\end{figure}
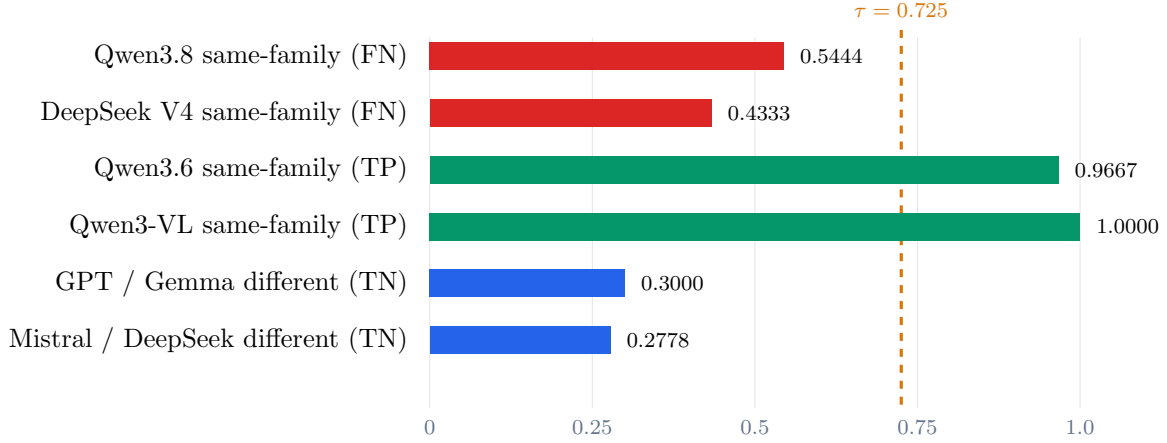

\begin{table}[htbp]
\centering
\small
\begin{tabular}{lccc}
\toprule
Metric & Estimate & Numerator/denominator & 95\% interval\\
\midrule
Eligible-pair coverage & 0.50 & 6/12 & ---\\
Balanced accuracy & 0.75 & 6 eligible pairs & ---\\
Sensitivity & 0.50 & 2/4 & 0.15--0.85\\
Specificity & 1.00 & 2/2 & 0.342--1.00\\
\bottomrule
\end{tabular}
\caption{Frozen-threshold holdout results. Metrics condition on strict pair
eligibility and must be read together with 50\% coverage.}
\label{tab:metrics}
\end{table}

\subsection{Why half of holdout is ineligible}

The validity contract prevents operational failures from becoming false evidence.
Table~\ref{tab:excluded} lists all excluded holdout pairs.

\begin{table}[htbp]
\centering
\small
\begin{tabular}{>{\raggedright\arraybackslash}p{3.7cm}>{\centering\arraybackslash}p{2.0cm}>{\raggedright\arraybackslash}p{7.1cm}}
\toprule
Pair & Informative repeats & Pre-specified exclusion reason\\
\midrule
GLM 5.3 / GLM 5.3 Flash & 1/3 & GLM 5.3 requests exceeded account-level RPM limits (HTTP 429).\\
Mistral Medium 3.5 / 3.1 & 0/3 & Mistral Medium 3.5 requests exceeded account-level RPM limits (429).\\
Qwen3.8 / DeepSeek V4 Vision & 0/3 & No endpoint matched the account data-policy/guardrail configuration (404).\\
Gemini 3.7 / GLM 5.3 Flash & 0/3 & Gemini successes omitted usage; additional requests were rate-limited.\\
Claude Haiku 4.5 / Qwen3.5 & 1/3 & Claude requests exceeded account-level RPM limits (429).\\
GLM 4.7 / Gemini 3.1 Lite & 0/3 & Gemini returned HTTP 200 for all requests but omitted usage in every response.\\
\bottomrule
\end{tabular}
\caption{Holdout exclusions. None is treated as a dissimilarity score.}
\label{tab:excluded}
\end{table}

Across holdout, all 2,160 formal logs are replayable and there are no transport
exceptions.  There are 189 non-200 responses and 157 HTTP-200 responses without
\texttt{usage.prompt\_tokens}.  Thus catalog availability does not imply that a
token-count fingerprint is observable for a specific account and route.

\section{Analysis}

\subsection{What the signal actually measures}

Equation~\ref{eq:score} detects whether two endpoints expose the same per-text count
sequence after a single translation.  A high value is compelling evidence of a
shared \emph{tokenization and request-wrapping stack}.  It is not direct evidence of
weight ancestry.  A stable hidden system prefix can create a constant offset;
provider accounting or chat templates can destroy one even if tokenizers share a
vocabulary.

The holdout false negatives make the distinction concrete.  Qwen3.8 Flash versus
Qwen3.8 27B obtains a mean of 0.5444, and dated DeepSeek V4 Flash versus V4 Pro
obtains 0.4333.  Their catalog names support same-family operational labels, yet
their count offsets vary substantially by text.  Possible causes include tokenizer
changes, multimodal versus text request wrapping, template differences, system
overhead, or provider routing.  The experiment cannot identify which mechanism is
responsible, and the paper does not infer one.

\subsection{Why the development result misleads}

The development set creates a clean gap: its lowest positive mean is 0.7889 and its
highest negative is 0.5333.  A threshold fit to this gap appears decisive.  The
holdout introduces new within-family transformations that are absent from
development.  The distribution shift is therefore conceptual, not merely
statistical: ``same family'' is a broader relation than ``same tokenization stack.''
More development pairs alone would not fix a mismatched target unless they expose
this heterogeneity.

\subsection{Operational coverage is part of validity}

A fingerprinting paper can overstate performance by scoring blocked or missing
observations as model failures.  Our strict treatment reduces eligible coverage to
50\%, but preserves semantic correctness: zero means observed disagreement, while
null means no valid comparison.  Reporting conditional accuracy without coverage
would hide half of the operational result.

\section{Limitations}
\label{sec:limitations}

\begin{enumerate}
  \item \textbf{Operational labels, not weight lineage.} Same-family labels come
  from provider namespaces and public version names.  They are useful for auditing
  product claims but do not prove shared training ancestry.
  \item \textbf{Small eligible holdout.} Strict validity leaves four positives and
  two negatives.  The Wilson intervals are consequently wide; specificity 1.0 is
  not strong evidence of a low population false-positive rate.
  \item \textbf{One gateway and one account.} Rate limits, usage omission, routing,
  and policy are account- and time-dependent.  Direct-provider APIs may have
  different coverage.
  \item \textbf{Thirty texts.} The corpus is heterogeneous but small.  It does not
  estimate language- or domain-specific performance and was not sampled from a
  defined population of text.
  \item \textbf{No adversarial spoofing.} A gateway can rewrite usage metadata or
  emulate another count sequence.  The method is diagnostic, not tamper-resistant.
  \item \textbf{No causal decomposition.} Tokenizer, template, hidden prompt, and
  accounting effects are entangled.  The score cannot attribute a mismatch to one
  layer.
  \item \textbf{Holdout cannot be reused for proof.} The observed failures may
  motivate a new method, but any revised feature or threshold requires a new
  untouched holdout.
\end{enumerate}

\section{Reproducibility, Data, and Ethics}

Code and sanitized data are publicly available at
\url{https://github.com/ictchenbo/which-llm}.  The repository contains the frozen
pair manifest, 30-text corpus, evaluator, validity contract, runner, calibration
script, analysis script, per-repeat JSON artifacts, and replayable JSONL logs.  The
main reproducibility artifacts are:
\begin{itemize}
  \item \texttt{experiments-v5.yaml}: 24-pair split and operational labels;
  \item \texttt{results/v5/tokenizer-calibration-frozen.json}: threshold and frozen
  digest;
  \item \texttt{results/v5/tokenizer-holdout-evaluation-final.json}: one-shot
  holdout evaluation;
  \item \texttt{results/v5/tokenizer-study-summary.json}: independent aggregate and
  call-health audit;
  \item \texttt{results/v5/*/repeat\_*.json} and \texttt{results/v5/calls/*.jsonl}:
  raw counts and call records.
\end{itemize}
The public data release removes authorization values, request and response headers,
call IDs, raw generated text, pilot runs, and the interrupted orphan log.  It retains
only fields required to replay token-count scoring and failure classification.  The
study uses ordinary API calls, does not attempt to bypass access controls, and
incurs provider charges.  We stopped before the planned full multi-dimensional
campaign because the primary research hypothesis failed and further spending would
not repair target mismatch.

\section{Conclusion}

Prompt-token counts are a useful black-box fingerprint when the claim is narrow:
two endpoints expose a highly consistent tokenization and request-wrapping stack.
They are not a reliable standalone test of named model-family membership.  A
perfect development result did not generalize: strict holdout coverage was 50\%,
and sensitivity among eligible pairs was 50\%.  The more general lesson is
methodological.  Black-box attribution studies should freeze thresholds, preserve
an untouched holdout, repeat observations over time, report observability coverage,
and distinguish an uninformative measurement from observed dissimilarity.  Under
that protocol, a negative holdout result is not a failed experiment; it is the
evidence that prevents an attractive proxy from becoming an unsupported lineage
claim.

\appendix
\section{Pair Matrix}
\label{app:pairs}

\footnotesize
\begin{longtable}{p{0.8cm}p{5.35cm}p{5.35cm}p{1.8cm}}
\toprule
ID & Model A & Model B & Label\\
\midrule
\endhead
d01 & \nolinkurl{z-ai/glm-5.3-flash} & \nolinkurl{z-ai/glm-5.2} & same\\
d02 & \nolinkurl{qwen/qwen3-235b-a22b} & \nolinkurl{qwen/qwen3-30b-a3b} & same\\
d03 & \nolinkurl{qwen/qwen3.5-9b} & \nolinkurl{qwen/qwen3.5-35b-a3b} & same\\
d04 & \nolinkurl{google/gemma-4-26b-a4b-it} & \nolinkurl{google/gemma-4-31b-it} & same\\
d05 & \nolinkurl{mistralai/ministral-3b-2512} & \nolinkurl{mistralai/ministral-8b-2512} & same\\
d06 & \nolinkurl{deepseek/deepseek-v4-flash} & \nolinkurl{deepseek/deepseek-v4-pro} & same\\
d07 & \nolinkurl{z-ai/glm-5.2} & \nolinkurl{deepseek/deepseek-chat} & different\\
d08 & \nolinkurl{qwen/qwen3.5-9b} & \nolinkurl{z-ai/glm-4.7-flash} & different\\
d09 & \nolinkurl{google/gemma-4-26b-a4b-it} & \nolinkurl{mistralai/mistral-small-2603} & different\\
d10 & \nolinkurl{deepseek/deepseek-v3.2} & \nolinkurl{qwen/qwen3.5-27b} & different\\
d11 & \nolinkurl{mistralai/ministral-8b-2512} & \nolinkurl{qwen/qwen3.5-9b} & different\\
d12 & \nolinkurl{z-ai/glm-5.3-flash} & \nolinkurl{google/gemma-4-31b-it} & different\\
\midrule
h01 & \nolinkurl{qwen/qwen3.8-flash} & \nolinkurl{qwen/qwen3.8-27b} & same\\
h02 & \nolinkurl{z-ai/glm-5.3} & \nolinkurl{z-ai/glm-5.3-flash} & same\\
h03 & \nolinkurl{deepseek/deepseek-v4-flash-0731} & \nolinkurl{deepseek/deepseek-v4-pro-0813} & same\\
h04 & \nolinkurl{qwen/qwen3.6-35b-a3b} & \nolinkurl{qwen/qwen3.6-27b} & same\\
h05 & \nolinkurl{qwen/qwen3-vl-30b-a3b-thinking} & \nolinkurl{qwen/qwen3-vl-30b-a3b-instruct} & same\\
h06 & \nolinkurl{mistralai/mistral-medium-3-5} & \nolinkurl{mistralai/mistral-medium-3.1} & same\\
h07 & \nolinkurl{qwen/qwen3.8-flash} & \texttt{deepseek/\allowbreak deepseek-v4-\allowbreak flash-\allowbreak vision-\allowbreak exp} & different\\
h08 & \nolinkurl{google/gemini-3.7-flash} & \nolinkurl{z-ai/glm-5.3-flash} & different\\
h09 & \nolinkurl{openai/gpt-5.4-nano} & \nolinkurl{google/gemma-4-26b-a4b-it} & different\\
h10 & \nolinkurl{anthropic/claude-haiku-4.5} & \nolinkurl{qwen/qwen3.5-9b} & different\\
h11 & \nolinkurl{mistralai/mistral-small-2603} & \nolinkurl{deepseek/deepseek-v3.2} & different\\
h12 & \nolinkurl{z-ai/glm-4.7-flash} & \nolinkurl{google/gemini-3.1-flash-lite} & different\\
\bottomrule
\end{longtable}

\section{Reproduction Commands}

\begin{lstlisting}
# offline regression tests
python3 test_fixes.py

# validate current catalog and pair balance
python3 validate_manifest_v5.py --manifest experiments-v5.yaml

# run tokenizer study by role (paid API calls)
python3 experiment_v5.py --manifest experiments-v5.yaml \
  --role development --all --repeats 3 --dims tokenizer --output results/v5
python3 experiment_v5.py --manifest experiments-v5.yaml \
  --role holdout --all --repeats 3 --dims tokenizer --output results/v5

# one-shot frozen-threshold evaluation
python3 calibrate_v5.py --results results/v5 \
  --frozen-calibration results/v5/tokenizer-calibration-frozen.json \
  --min-informative-repeats 3 \
  --output results/v5/tokenizer-holdout-evaluation-final.json

# independent study summary
python3 analyze_tokenizer_v5.py
\end{lstlisting}


\begin{thebibliography}{99}

\bibitem{Jin2024}
H.~Jin, C.~Zhang, S.~Shi, W.~Lou, and Y.~T.~Hou.
ProFLingo: A Fingerprinting-based Intellectual Property Protection Scheme for Large
Language Models. \emph{arXiv:2405.02466}, 2024.

\bibitem{YangWu2024}
Z.~Yang and H.~Wu.
A Fingerprint for Large Language Models.
\emph{arXiv:2407.01235}, 2024; revised 2025.

\bibitem{Iourovitski2024}
D.~Iourovitski, S.~Sharma, and R.~Talwar.
Hide and Seek: Fingerprinting Large Language Models with Evolutionary Learning.
\emph{arXiv:2408.02871}, 2024.

\bibitem{Yang2025}
Z.~Yang, Y.~Wu, Y.~Shen, W.~Dai, M.~Backes, and Y.~Zhang.
The Challenge of Identifying the Origin of Black-Box Large Language Models.
\emph{arXiv:2503.04332}, 2025.

\bibitem{Hu2026}
Y.~Hu, Z.~Jiang, M.~Li, O.~Ahmed, Z.~Huang, C.~Hong, and N.~Z.~Gong.
Fingerprinting LLMs via Prompt Injection.
In \emph{Proceedings of ACL 2026}, pages 11795--11810, 2026.

\bibitem{Fang2026}
Y.~Fang, Y.~Feng, B.~Li, and M.~Zhou.
KBF: Knowledge Boundary as Fingerprint for Language Model and Black-Box API
Auditing. \emph{arXiv:2605.29524}, 2026.

\bibitem{White2026}
I.~White, Y.~Jafari, and T.~Berg-Kirkpatrick.
Black-Box Forensics for Conversational LLM Agents.
\emph{arXiv:2606.22698}, 2026.

\bibitem{OpenRouterModels}
OpenRouter.
Models documentation: tokenization, usage, pricing, and architecture metadata.
\url{https://openrouter.ai/docs/guides/overview/models}. Accessed August 30, 2026.

\bibitem{OpenRouterModelsAPI}
OpenRouter.
Models API reference: \texttt{GET /api/v1/models}.
\url{https://openrouter.ai/docs/api/api-reference/models/get-models}.
Accessed August 30, 2026.

\end{thebibliography}
\end{document}